\documentclass[letterpaper, 10 pt, conference]{ieeeconf}  

\IEEEoverridecommandlockouts                              

\usepackage{graphics} 
\usepackage{epsfig} 
\usepackage{mathptmx} 
\usepackage{times} 
\usepackage{amsmath} 
\usepackage{amssymb}  
\usepackage[sort]{cite}
\usepackage{subfig}
\usepackage{hyperref}
\title{\LARGE \bf
Analytic Model for Quadruped Locomotion Task-Space Planning*
}

\author{Carlo~Tiseo,$^{1}$, Sethu Vijayakumar,$^{1}$,~Michael~Mistry$^{1}$ 
\thanks{*This research was supported by the EPSRCs RAI Hub for Extreme and Challenging Environments: ORCA EPR026173/1.}
\thanks{$^{1}$Carlo Tiseo, Sethu Vijayakumar and Michael Mistry are with the School of Informatics at the University of Edinburgh, Bayes Centre, 47 Potterrow, Edinburgh, EH8 9BT, UK, e-mail: ctiseo@ed.ac.uk, sethu.vijayakumar@ed.ac.uk, mmistry@ed.ac.uk}
}

\begin{document}

\pagestyle{empty}
\begin{table*}[t]
	IEEE Copyright Notice \textcopyright 2019 IEEE. Personal use of this material is permitted. Permission from IEEE must be obtained for all other uses, in any current or future media, including reprinting/republishing this material for advertising or promotional purposes, creating new collective works, for resale or redistribution to servers or lists, or reuse of any copyrighted component of this work in other works.\\
	
	Accepted to be Published in 2019, 41th Annual International Conference of the IEEE Engineering in Medicine and Biology Society (EMBC), Berlin Germany. 
\end{table*}

\newpage

\maketitle
\thispagestyle{empty}
\pagestyle{empty}

\begin{abstract}

Despite the extensive presence of the legged locomotion in animals, it is extremely challenging to be reproduced with robots. Legged locomotion is an dynamic task which benefits from a planning that takes advantage of the gravitational pull on the system. However, the computational cost of such optimization rapidly increases with the complexity of  kinematic structures, rendering impossible real-time deployment in unstructured environments. This paper proposes a simplified method that can generate desired centre of mass and feet trajectory for quadrupeds. The model describes a quadruped as two bipeds connected via their centres of mass, and it is based on the extension of an algebraic bipedal model that uses the topology of the gravitational attractor to describe bipedal locomotion strategies. The results show that the model generates trajectories that agrees with previous studies. The model will be deployed in the future as seed solution for whole-body trajectory optimization in the attempt to reduce the computational cost and obtain real-time planning of complex action in challenging environments.     

\end{abstract}

\section{INTRODUCTION}
The last decades have been characterised by a shift of paradigm in robotics from traditional industrial automation approaches towards a more human centric design that aims to improve both living and working conditions for humans. New robotic branches have formed to address issues regarding healthcare, haptics, human-robot interaction and collaboration have underlined the limited adaptability of traditional robotics, which was not conceived for dealing with interaction and uncertainties \cite{Xin2018}. Therefore, scientists have started looking at animals as source of inspiration due to their dynamic dexterity in their daily living \cite{Tiseo2018,Bucher2015,Tiseo2018a}. The study of how animals interact with environmental dynamics underlined the importance of having some degree of intrinsic softness in the mechanical structure, which lead to the development of soft-actuators (e.g., Variable Stiffness and Serial Elastic Actuators)\cite{Xin2018}. Nevertheless, despite the introduction of these technologies have improved the performances of our robots in reacting to external perturbation, they are not comparable with animals' abilities.

The identification of efficient and effective planning strategies in unstructured environments is critical to the development of mobile robotics platform that can enter daily living environments to improve human life quality and take over dangerous tasks \cite{Tiseo2018,Xin2018}. Quadruped robotics is among the viable solutions to replace humans operators in hazardous environments (e.g., deep mines, offshore oil rigs and disaster sites), offering both greater flexibility than wheeled robots and better stability characteristics compared to bipeds \cite{Xin2018}. However, the high redundancy of the system kinematics implies a higher complexity in the planning stage, making the system less adaptable to sudden changes in the environment which trigger a replanning of the entire action. Yet again, we observe that animals are better than robots in reacting to perturbations, despite the limitations of their neural system in both transmission and elaboration of the information \cite{Tiseo2018, Tommasino2017, Minassian}. A possible answer to how nature could adapt to compensate for its intrinsic limitations may come to the latest studies in human motor control. Particularly interesting is the concept of dynamic primitives. They are collection of basic movements built through the interaction with external attractors, which can be combined to perform complex actions. This theory is an extension of the motor primitives that considered only the internal dynamics of the human body. The latter has also been integrated in a forward controller called the Passive Motion Paradigm (PMP) that can accurately reproduce reaching planar movements and wrist pointing task, where the external dynamics can be neglected \cite{Tiseo2018, Tiseo2018a, Tommasino2017}. 
\begin{figure*}[thpb]
	\centering
	\includegraphics[scale=0.65]{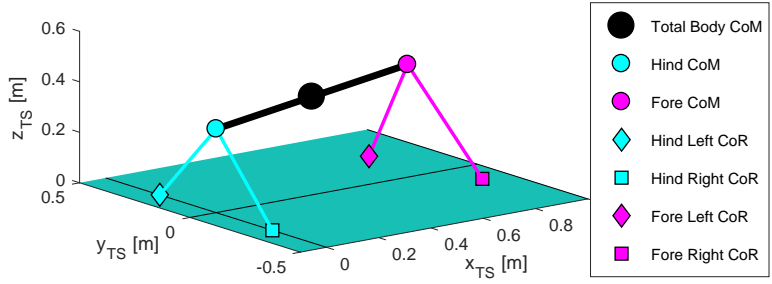}   
	\caption{The simplified quadrupedal model is composed by two bipeds connected by a rigid spine. Each biped is fully described via the Cartesian coordinates of its Centre of Mass (i.e., Hind/Fore CoM) and the Centre of Rotations (i.e., Fore/Hind Left/Right CoR).}
	\label{fig1}
\end{figure*}

Tommasino \textit{et al} have recently formulated an extension to the passive motion paradigm, called $\lambda_{dyn}-PMP$, which extend the theory to include both the motor synergies and the interaction with an external environment \cite{Tommasino2017}. Meanwhile, Tiseo \textit{et al} have uncovered evidence that the $\lambda_{dyn}-PMP$ model is also applicable to human locomotion strategies, which they have shown to be a dynamic reaching task on the gravitational attractor acting on the Centre of Mass (CoM) \cite{Tiseo2018}. In particular, their results show that the spatio-temporal gait parameters (i.e., step length, step width, velocity and step frequency) are correlated by well defined relationship to achieve a stable behaviour that take advantage of the intrinsic dynamics of the biped. Thus, describing human locomotion as a dynamic reaching task that deploys the fixed points of the gravitational attractor to produce the close cycle behaviour that we observe in humans. The scope of this paper is to extend the model proposed in \cite{Tiseo2018} to quadrupedal robots by modelling them as two synchronised bipeds with the two masses connected by a rigid rod as proposed in \cite{Griffin2004}. The aforementioned behaviour is also observed in humans when two individuals are carrying a ladder as reported in \cite{Lanini2017}.   

\section{Methods and Materials}
The section presents with the extension of the bipedal model proposed by Tiseo \textit{et al} in \cite{Tiseo2018}, and the validation method used to test the model for quadrupeds.

\subsection{Extension of the Bipedal Model}
The gravitational attractor model of locomotion is based on a linear inverted pendulum model with variable length described in the task-space. It relies on the analysis on the intrinsic dynamics of the mechanical system to determine the relationships within the spatio-temporal parameters of the gait and the trajectories for the CoM and the foot swing. The extension to the quadruped is done considering two bipeds connected as shown in Fig. \ref{fig1}. The following formulation also considers that the feet do not move laterally; thus maintaining their lateral coordinates equal to half the step width ($\pm d_{SW}/2$).
\begin{equation}
\label{eq1}
\begin{array}{l}
\vec{x}_{CoMi}(t)=\left\{\begin{array}{l}
x_{CoMi}=v_w t+x_{0\text{-}CoMi}\\
y_{CoMi}=A_y\cos(\pi \omega_{S}t+\phi_{i})\\
z_{CoMi}=(z_{max}-Az)+A_z\cos(2\pi \omega_{S}t+2\phi_{i})
\end{array}\right.\\
x_{CoRij}(t)=\left\{\begin{array}{lr}
x_{0\text{-}CoRij} & \textit{Support (SU)}\\
\frac{y_{CoR\text{-}S}-y_{CoR\text{-}SU}}{m_{CoM|CoR\text{-}SU}}+x_{CoR\text{-}SUi}& \textit{Swing (S)}
\end{array}\right.\\
z_{CoRij}(t)=\left\{\begin{array}{lr}
0 & \textit{SU}\\
z_c cos(\omega_{S}t+\phi_{i})& \textit{S}
\end{array}\right. 
\end{array}
\end{equation}
where i=F,H (Fore, Hind) indicates the fore or hind biped, j=R,L (Right, Left) refers to the foot. $l$ is the pendulum length, $d_{SL}$ is the step length, $v_w$ is the walking velocity, and $\omega_S$ is the step frequency (i.e., cadence), $z_{max}=\sqrt{l^2-(d_{SW}/2-Ay)^2}$ is the maximum height of CoM, and $\phi_i$ is the starting phase. $A_y=d_{SW}/(2\pi\omega_{S}d_{SL})=d_{SW}/(2\pi v_w)$ is the mediolateral amplitude of the CoM trajectory, and $A_z=z_{max}-\sqrt{L^2-(d_{SW}/2)^2-(d_{SL})^2}$ is the vertical amplitude of the CoM trajectory in which L describes the maximum extension achievable by the leg. $x_{0\text{-}CoMi}$ and $x_{0\text{-}CoRij}$ are the initial coordinates for the CoM and the Centre of Rotation (CoR), respectively. $m_{CoM|CoR\text{-}S}$ is the slope of the segment connecting the support foot CoR to the CoM and $x_{CoR\text{-}Si}$ is the x-coordinate of the foot providing support. $z_c$ is the desired vertical clearance during swing, that has currently been set at 5 cm. The CoM trajectory of the quadruped is obtained from the bipeds' CoM trajectories as follows:
\begin{equation}
\label{eq2}
\begin{array}{l}
\vec{x}_{CoM}(t)=\frac{\vec{x}_{CoMF}+\vec{x}_{CoMH}}{2}
\end{array}
\end{equation}
\begin{figure*}[h]
	\centering
	\subfloat[][]{\includegraphics[width=0.8\columnwidth]{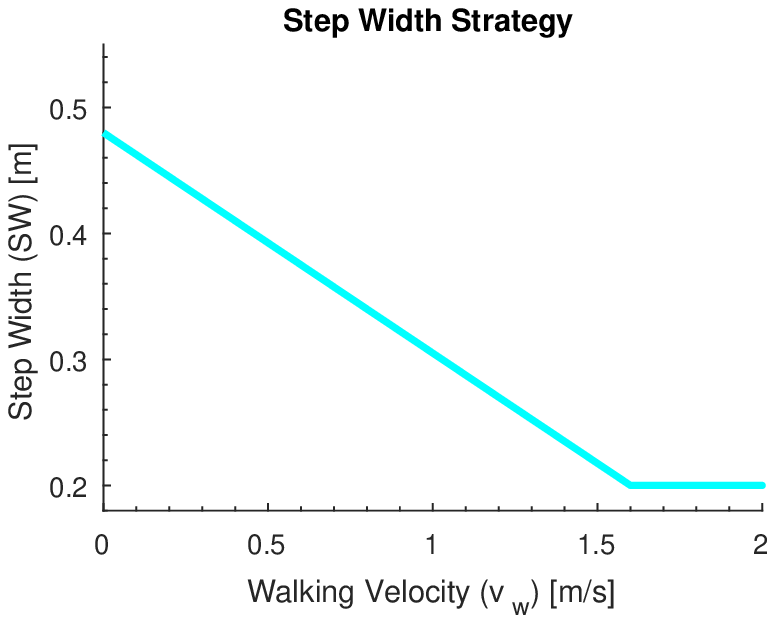}}
	\hfil
	\subfloat[][]{\includegraphics[width=0.8\columnwidth]{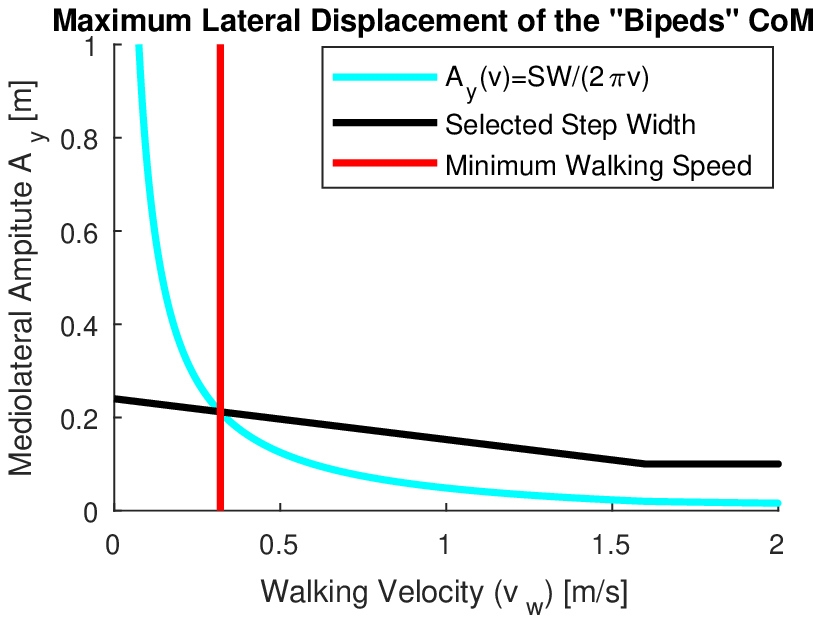}}
	\hfil
	\subfloat[][]{\includegraphics[width=0.8\columnwidth]{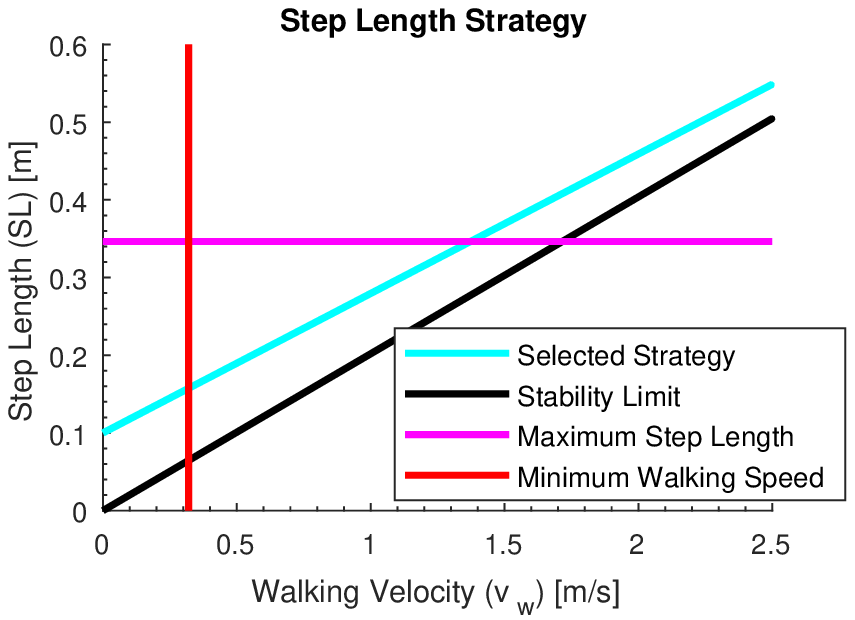}}
	\hfil
	\subfloat[][]{\includegraphics[width=0.8\columnwidth]{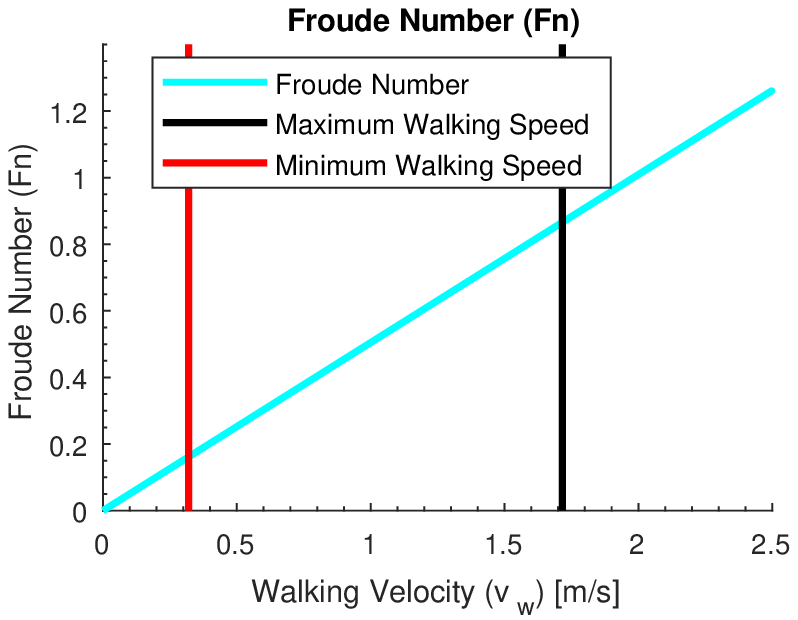}}
	\hfil
	\caption{(a) The SW strategy as defined in eq. (\ref{eq3}). (b) The figure shows that the selection of the SW determines the minimum walking velocity for the system. (c) The SL strategy as defined in eq. (\ref{eq4}). However, stable strategies have to be included in the area delimited between the maximum reachable distance and the minimum SL determined by the natural frequency of the pendulum ($\omega_n$). (d) The Froude Numbers associated with the range of velocities determined by the proposed model.}
	\label{fig2}
\end{figure*} 
\subsection{Gait Parameters}
The model requires to define the gait parameters as function of the walking velocity to generate the CoM and CoR trajectories in equations (\ref{eq1}) and (\ref{eq2}). The Step Width (SW) has been defined based on constraints determined from previous experiment conducted with the ANYmal robot \cite{Xin2018}. 
\begin{equation}
\label{eq3}
\begin{array}{l}
d_{SW}=\left\{\begin{array}{lc}
 m_{sw}v_w+d_{SWmax}, & \textit{if $d_{SW}$ $\ge$ $d_{SWmin}$}\\
 d_{SWmin}, & \textit{otherwise}
\end{array}
\right. \\
\end{array}
\end{equation}
$$
\begin{array}{lll}
            m_{sw}=\frac{d_{SWmin}-d_{SWmax}}{2},&
            d_{SWmax}=1.2D_{ML},& d_{SWmin}=\frac{D_{ML}}{2}
\end{array}
$$
where $D_{ML}$ is the mediolateral distance between the hip joints. The choice of the SW also introduces a limit of the minimum speed where the ballistic trajectory of the CoM is constrained within the two feet. Despite the impossibility to fully exploit the intrinsic inverted pendulum dynamics at lower speed, it is still possible to walk by constraining the CoM within the feet and actively track the ballistic trajectories during the foot transitions. Differently for the SW, the Step Length (SL) strategy accounts for the limitation imposed by the inverted pendulum dynamics, and it is defined as follows:

\begin{equation}
\label{eq4}
\begin{array}{l}
d_{SL}=m_{SL}v_w+d_{SLmin}\\
\end{array}
\end{equation}
$$
 \begin{array}{ll}
 m_{SL}=\frac{d_{SLmax}-d_{SLmin}}{0.8v_{w\text{-}max}},& \omega_n=\sqrt{\frac{g}{l}}, \text{ }v_{w\text{-}max}=\omega_nd_{SLmax},\\
 d_{SLmin}=0.1 [m],&d_{SLmax}=0.35 [m]\\

 \end{array}
$$
A Matlab (MathWoks, US) code was implemented for the simulations and only two inputs are $v_w$ and number of steps. The gait constraint identified by the model are compared with animal data to verify their consistency, and using the Froude Number (Fn=$v_w^2/(gl)$) to classify the gait strategies \cite{Jayes1978}. Sequentiality, the trajectories obtained with our model are compared with the results from previous research that analysed the coupling of two bipeds using either a rigid or a \textit{quasi}-rigid interface \cite{Lanini2017,Griffin2004}.
\begin{figure*}[thpb]
	\centering
	\subfloat[][]{\includegraphics[width=0.85\columnwidth]{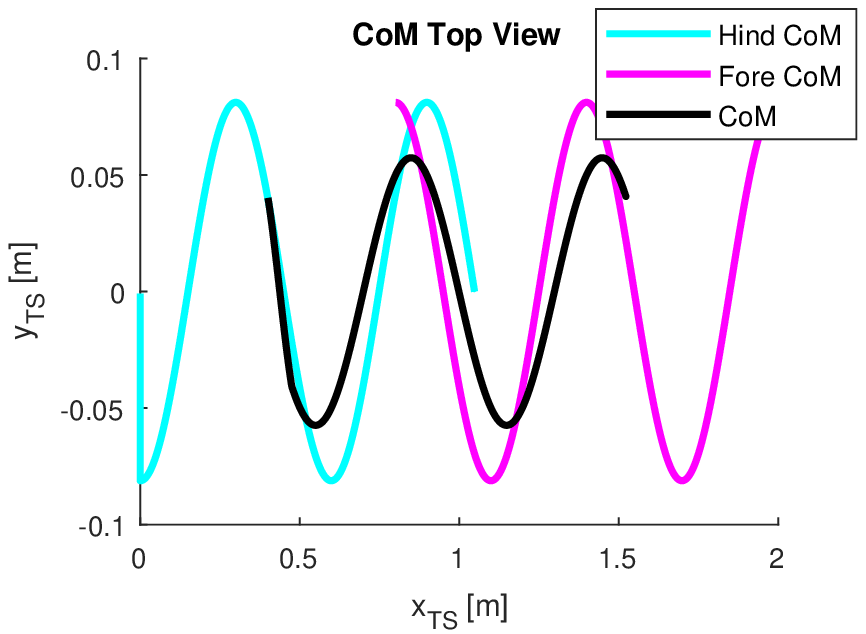}}
	\hfil
	\subfloat[][]{\includegraphics[width=0.85\columnwidth]{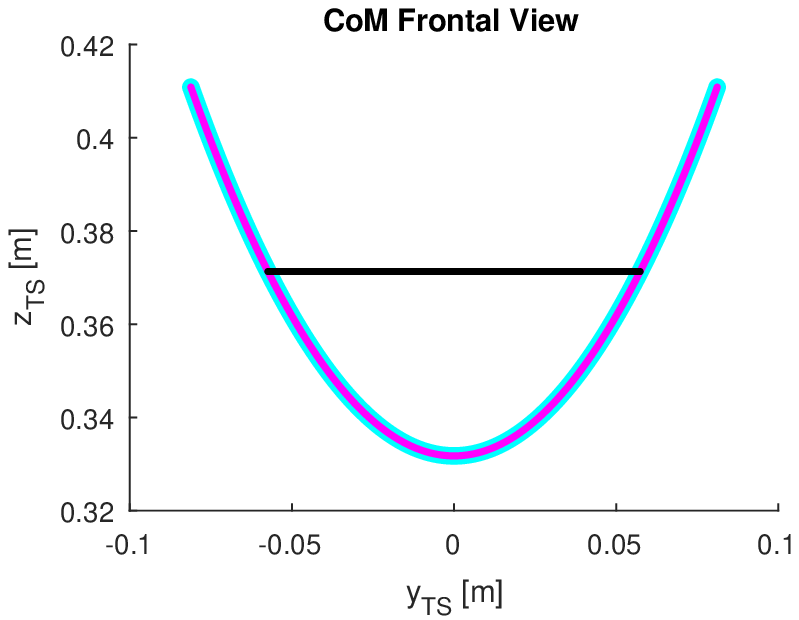}}
	\hfil
	\subfloat[][]{\includegraphics[width=0.85\columnwidth]{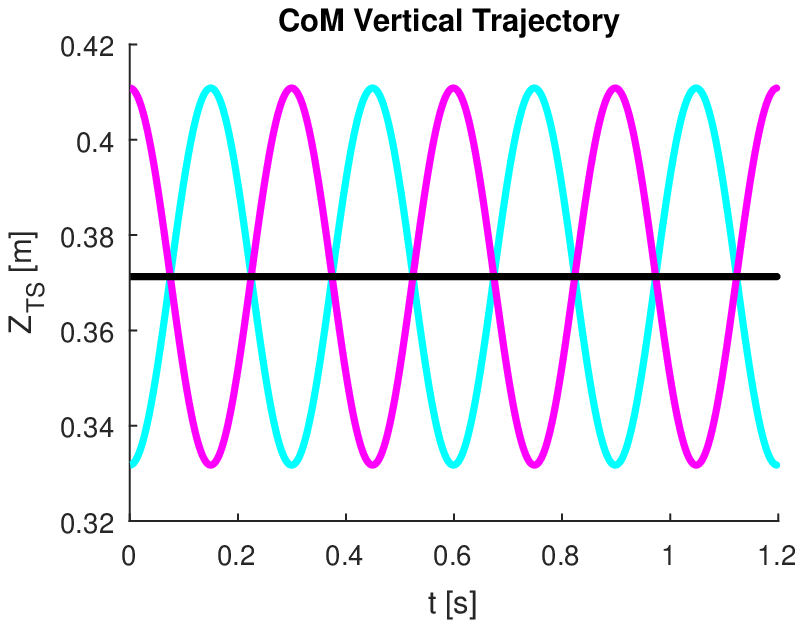}}
	\hfil
	\subfloat[][]{\includegraphics[width=0.85\columnwidth]{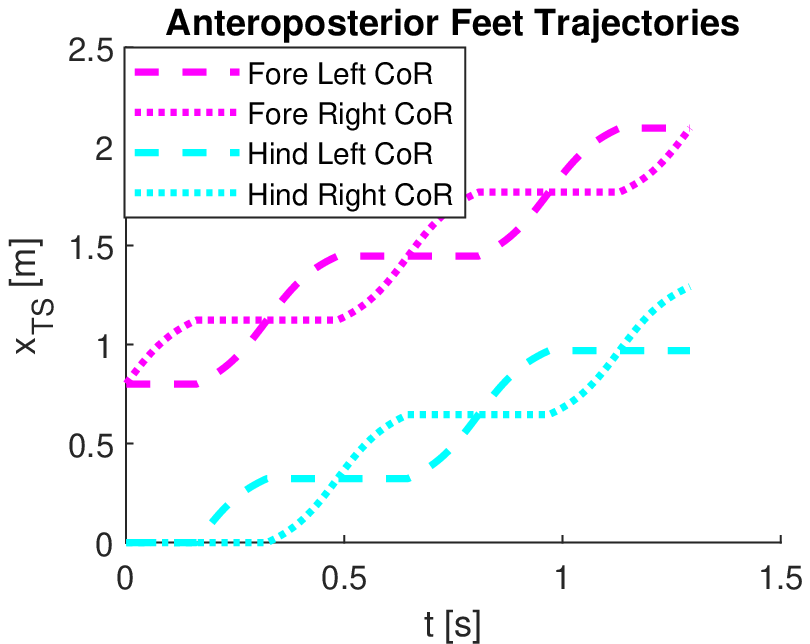}}
	\hfil
	\caption{Sample trajectories generated at 1 [m/s]. (a) The trajectory of the CoMs on the transverse plane. (b) The trajectory of the CoMs on the frontal plane. (d) The trajectory of the CoMs on the lateral plane, which reproduces results similar to both \cite{Griffin2004} and \cite{Lanini2017}. (c) Anterior Posterior Trajectories for the 4 CoRs.}
	\label{fig3}
\end{figure*}
\begin{figure*}[!t]
	\centering
	\includegraphics[scale=0.85]{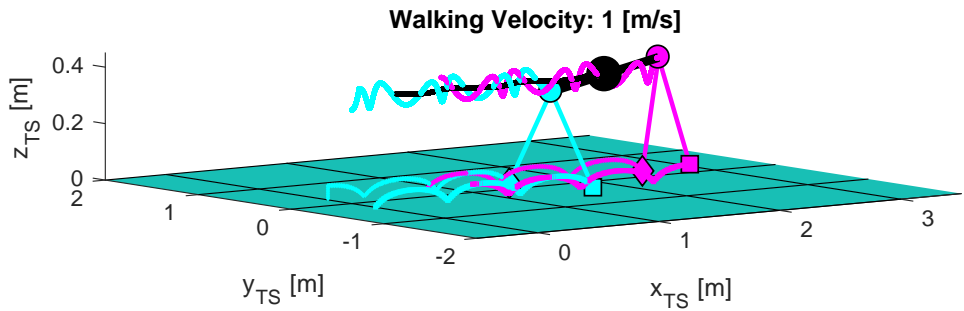}   
	\caption{A sample walking trajectory obtained at 1 [m/s] with the proposed model. It uses the graphic conventions defined in Fig. \ref{fig1}, and the axes are the task-space coordinates. The video is available at  \href{https://www.youtube.com/watch?v=6wsxPQk7zLk\&feature=youtu.be}{https://www.youtube.com/watch?v=6wsxPQk7zLk\&feature=youtu.be} .}
	\label{fig4}
\end{figure*}
\section{Result}
The range of walking speed admissible by our model is between 0.32 and 1.71 [m/s], which based on the Froude Number classification ranges from walking to trot. Specifically, it belongs to the walking range of dogs (Fn=[0.10,0.40), \cite{Jayes1978}) until a speed of 0.80 [m/s], before transitioning to the gait/canter range (Fn=[0.40, 4.00], \cite{Jayes1978}, as reported in Fig. \ref{fig2}. The results show that the walking range is almost fully covered being the Froude Number at the minimum speed equal to 0.16; meanwhile, the maximum Fn achieved by the proposed model is 0.86. However, it shall be considered that animal trot includes a flight phase at higher speed, while our model imposes the absence of such phase. The trajectories generated by our model, shown in Fig. \ref{fig3} for a walking speed of 1.00 [m/s], are consistent with both the results obtained by previous research both for quadruped and coupled humans.    
\section{Discussion}
The data confirm that the model proposed by Tiseo \textit{et al.}, \cite{Tiseo2018}, can be extended to quadrupedal locomotion, and it can predict a range of admissible speed in dogs. These preliminary results show the theoretical feasibility of computationally efficient planner for quadrupedal based on the exploration of the gravitational dynamics. The proposed approach may also be used to reduce the computational cost of current methods by limiting their solution space to the set of solutions compatible with the saddle attractor locomotion theory. Furthermore, the model extension to quadrupeds shows interesting parallel with bipeds motor control, which may help to understand how the brain controls locomotion via the modulation of the Central Pattern Generators (CPG) located in the brain stem and spinal cord \cite{Minassian,Bucher2015}. 

In conclusion, the extension of the model to quadrupedal locomotion has been proved possible but, there is still work to be done before it can tested on real scenarios and perform essential locomotor behaviours (e.g., uneven terrain locomotion, turning, and obstacle avoidance). Nevertheless, we were able to gain some essential knowledge on how the body structure of a robot limits its locomotion abilities, that can be useful for the improvement of the hardware design.

\addtolength{\textheight}{-0cm}   








\end{document}